# Emerging trends in Cislunar Space for Lunar Science Exploration and Space Robotics aiding Human Spaceflight Safety

**Arsalan Muhammad[1,2]\*, Yue Wang[1], Hai Huang[1,2], Hao Wang[2]**

[1]*School of Astronautics, Beihang University*
*Xueyuan Road, 102206, Beijing, China*
[2] *Regional Centre for Space Science and Technology Education in Asia & the Pacific (Affiliated to the United Nations) (RCSSTEAP), Hangzhou International Innovation Institute of Beihang University, China;*
\* Corresponding Author

**Abstract**

In recent years, the Moon has emerged as an unparalleled extraterrestrial testbed for advancing cutting-edge technological and scientific research critical to enabling sustained human presence on its surface and supporting future interplanetary exploration. This study identifies and investigates two pivotal research domains with substantial transformative potential for accelerating humanity's interplanetary aspirations: First is Lunar Science Exploration with Artificial Intelligence and Space Robotics which focusses on AI and Space Robotics redefining the frontiers of space exploration. As humanity advances toward sustained presence on the Moon, these technologies are becoming indispensable tools for navigating, analyzing, and utilizing the lunar environment with unprecedented precision and adaptability. Second being Space Robotics aid in manned spaceflight to the Moon serving as critical assets for pre-deployment infrastructure development, In-Situ Resource Utilization (ISRU), surface operations support, and astronaut safety assurance. By integrating autonomy, machine learning, and real-time sensor fusion, space robotics not only augment human capabilities but also serve as force multipliers in achieving sustainable lunar exploration, paving the way for future crewed missions to Mars and beyond.

## 1. Introduction

With the swift progression of space technologies, there has been a growing emphasis on long-duration and deep-space planetary missions, expanding the frontiers of extraterrestrial exploration [1]. The distinctive characteristics of the lunar space environment offer groundbreaking opportunities for the future of space exploration. By enabling the development and sustainable use of the Moon's surface, these conditions can support extended human presence, facilitate deeper space missions, and contribute to the long-term advancement and sustainability of human civilization [2][3]. This paper aims to provide the insights related to Lunar Science Exploration with Artificial Intelligence and Space Robotics field along with the Space robotics aiding in manned spaceflight missions to the Moon and beyond.

## 2. Methodology

This review is intended to serve as a valuable resource for future scientists in advancing their research, while also supporting engineers and academic professionals in designing and conducting laboratory experiments, as well as developing intelligent algorithms tailored to the field of lunar science and exploration. Two major research domains focusing on the subject area have been explored along with their key research questions, as shown in table 1

**Table 1:** Key Research Domains related to Lunar Science Exploration

| Key research domain | Topic Question |
|---|---|
| Lunar Science Exploration with Artificial Intelligence and Space Robotics | 1. What role can autonomous robotics and AI play in lunar surface mapping and habitat construction?<br>2. How can AI-driven decision-making enhance the efficiency and adaptability of lunar exploration missions?<br>3. How the human-robot cooperation can improve in-situ resource utilization (ISRU) on the Moon? |
| Space Robotics aid in manned spaceflight to the Moon | 1. How legitimate is the space travel to humanity? What are the key trade-offs between human and robotic exploration in terms of mission effectiveness, cost, and risk for lunar operations? |
| | 2. How can AI-augmented robotics enhance astronaut safety and productivity during surface operations on the Moon? |
| | 3. What ethical, legal, and operational frameworks are needed to manage risk and responsibility in crewed lunar exploration? |

## 3. Results

For both the key research domains, substantial number of papers have been reviewed from well-known sources and databases. Table 2 shows the summary of year-wise distribution of selected articles with their database while figure 1 shows the graphical representation of this summary.

**Table 2:** Year-wise & Database source-wise distribution of selected articles in each identified key research domain

| Key research domain | Years | Database Source | Total |
|---|---|---|---|
| Lunar Science Exploration with Artificial Intelligence and Space Robotics | 2024 (3), 2023 (3), 2021 (2), 2020 (2), 2019 (1), 2014 (1), 2013 (1), 2012 (2), 2009 (2), 2006 (1), 2003 (1) | IEEE (3), ArXiv (5), ACM Digital Library (1), Google Scholar (1), Science Direct (2), PubMed (1), Springer (3), NPJ Microgravity (1), Reviews in Minerology and Geochemistry (1) | 19 |
| Space Robotics aiding in manned spaceflight to the Moon and beyond | 2025 (2), 2024 (2), 2023 (2), 2022 (3), 2021 (1), 2020 (3), 2019 (4), 2017 (3), 2016 (1), 2014 (1), 2012 (1), 2011 (1), 2006 (2), 2005 (1), 2003 (2), 2002 (1), 2001 (1), 1999 (2), 1992 (1), 1990 (1) | IEEE (3), Advancing Earth and Space Science (1), ArXiv (3), Aerospace Research Council (5), Science Direct (11), PubMed (7), MDPI (1), Springer (2), NPJ Microgravity (2) | 35 |

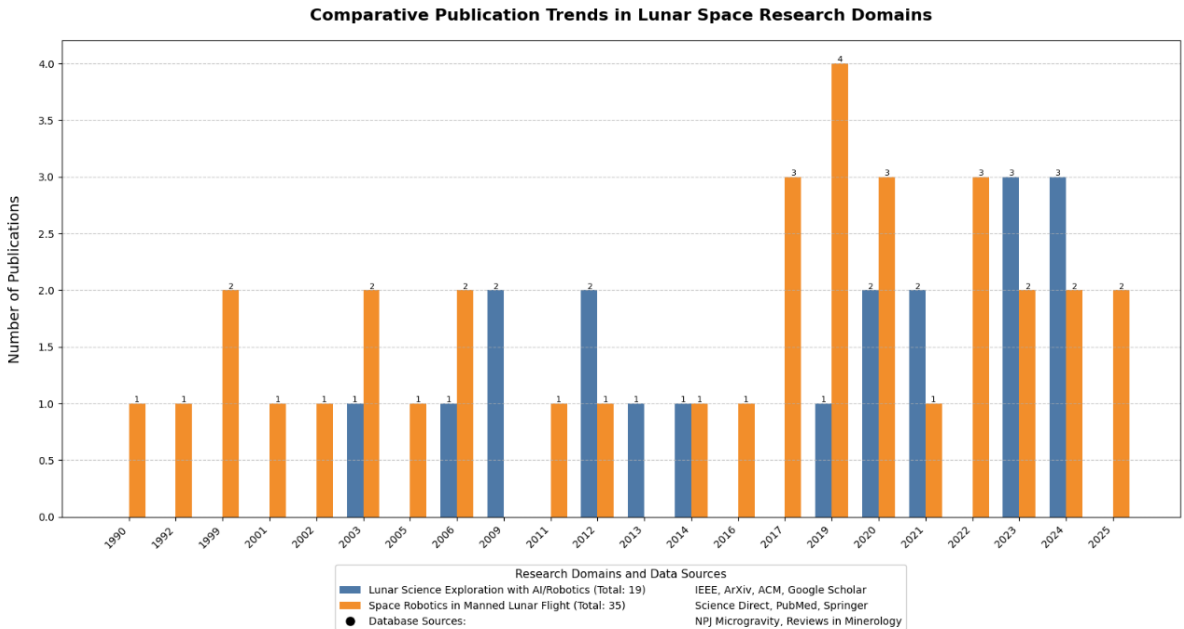

**Figure 1:** Year-wise publication trend

*3.1 Lunar Science Exploration with Artificial Intelligence and Space Robotics*

The Moon operates as a critical geological repository, preserving high-fidelity records of the early Solar System's evolutionary history. It encapsulates essential data related to planetary accretion, differentiation, and impact processes, as well as stellar evolution. Furthermore, it provides scientifically rich insights into fundamental astrophysical and Helio physical mechanisms, offering a unique window into deep space environments, cosmic ray interactions, and broader exoplanetary system dynamics (Franchini et al., 2024). The objective of lunar research investigations is to address fundamental questions regarding the Moon's formation and evolutionary history, as well as the geological processes it has undergone. These studies also aim to characterize the presence and distribution of water and other volatiles, analyze the tenuous lunar atmosphere, dust environment, and plasma interactions, and explore the Moon's potential as a strategic hub for human expansion. This includes assessing its role in enabling future space industrialization, resource utilization, and serving as a launchpad for sustained human missions to Mars and deeper space destinations beyond [4] [5].

Table 3 shows the research characterization in which the lunar science domain could be categorized for scientific and future exploration.

**Table 3:** Lunar Science exploration classification table

| Categorization | Research purpose | Fundamental scientific questions |
|---|---|---|
| Astro Physics and Scientific exploration | Scientific activities to address fundamental science questions about the evolution of Solar System, Universe, the next destination of Humans from Earth to Moon and beyond [6]. The Moon surface can also be investigated for biological investigations about life sciences for evolution and adaptation of organisms in space [7]. | 1. Understand the impact history of Solar system as recorded on the Moon and geological and environmental impacts of lunar exploration. 2. A platform for research between astrophysics, fundamental physics concepts of general relativity, highly energised cosmic rays etc |
| Moon surface for future missions to Mars and beyond | Lunar platform for scientific exploration further provides the key learning and technologies for risk reduction in future missions to Mars and beyond. | 1. Develop capabilities for efficient human-machine interaction, autonomous crew operations including health and safety for sustainable development. |
| AI based multi-rover collaborative systems for terrain mapping and sampling | The primary purpose is to develop scalable, AI-driven systems for coordinating multiple rovers to perform safe, efficient terrain mapping and geological sampling on the lunar surface, reducing collision risks. | 1. How can deep reinforcement learning optimize collaborative path planning to avoid obstacles and ensure safe inter-rover coordination in dynamic lunar terrains. |
| AI driven robotic construction and in-situ adaptation for scientific outposts. | This research aims to create intelligent, adaptive robots that autonomously construct and modify scientific outposts using 3D printing technologies and lunar regolith, promoting sustainability and self-sufficiency for long-term human presence. | 1. How can AI algorithms enable real-time adaptation of construction techniques to varying lunar soil compositions and environmental hazards. 2. How to integrate ISRU processes with robotic systems to ensure structural durability in vacuum conditions. |
| Lunar exploration though Virtual Reality simulations | Provision of a comprehensive research environment incorporating VR simulations to reproduce the opportunities and challenges of Moon exploration with aim for mission success and astronauts training [8][9][10] | 1. To perform lunar exploration and expertise are required from building and managing the in-situ infrastructure, precise PNT, human robot interaction, astronaut training. 2. To build a VR simulation environment capable of training the user in future Moon missions. |
| AI/ ML empowering lunar exploration | AI and ML based techniques such as convolutional/ deep space neural networks for; Remote mapping of lunar surface to perform in-depth analysis about lunar morphology, Lunar orbit spectroscopy for analysing surface minerology & Lunar surface exploration for landing site feasibility and rover driving technologies for planetary surface research [11][12] | 1. Understand about convolutional and deep space neural networks and integrating it with computer vision techniques for gaining advantages using AI/ ML based automation. |
| Space Robotics Revolutionization | Cooperation between humans and space robots for performing space based extra vehicular activities with limited resources (power, time) and harsh environmental conditions [13][14] | 1. Modelling the extreme conditions of unpredictable lunar environment for lunar surface operations including the uncertainties, time delays, low gravity. 2. Enhancing the reliability and autonomy of space robotic systems. 3. Development of sophisticated algorithms for decision making, improved sensor capability for real-time situational awareness. |

*3.2 Space Robotics aiding in manned spaceflight to the Moon and beyond*

Six decades after the dawn of human spaceflight, questions being asked, how legitimate is space travel for humanity? Upcoming missions to the Moon and Mars present unprecedented challenges, each carrying distinct risks and potential rewards. The incidents and accidents in manned space flight has been summarized by [15] as, out of 327 manned space flights from 1961 to 2020, 5 accidents and 36 incidents were reported resulting in 5 fatalities and 3 space craft losses. However, the total number of incidents decreased from 8 in 1960s to 3 till 2010. Moreover, no astronaut fatality has been reported since 2003 due to greater international cooperation and technological advancement [16]. While there are several challenges and risks associated while exploring the Moon's hazardous surface where the risk to astronaut safety lies at every step, and also the communication delays and frequent radio blackouts between Earth and the Moon will demand greater astronaut autonomy [17][18][19].

A persistent critique from the scientific community questions the necessity of human spaceflight, often dismissing it as a costly spectacle when robotic missions could achieve scientific objectives more efficiently. Proponents of this view argue that unmanned spacecraft offer superior cost-effectiveness and risk mitigation, particularly for planetary exploration. However, empirical studies demonstrate that human presence remains critical for tasks requiring real-time adaptability, such as deploying intricate instrumentation or conducting geological fieldwork activities where human dexterity and contextual decision-making outperform robotic capabilities [20][21]. There exists a well-defined distribution of tasks between humans and machine which are of evolutionary nature [22] which includes; installation and maintenance of sensitive equipment, humans as a field scientist, the concept of telepresence, science education and research.

Table 4 shows the research characterization in which space robotics aiding manned spaceflights to the Moon for scientific and future exploration.

**Table 4:** Classification table for Space Robotics aiding Manned Spaceflights to the Moon

| Categorization | Research purpose | Fundamental scientific questions |
|---|---|---|
| Scientific discovery & exploration | Human missions enable direct exploration beyond robotic capabilities allowing astronauts to make real-time decisions while conducting complex experiments. | 1. How can crewed missions integrate multi-disciplinary studies to address the Moon's role as a record of Earth's early history? |
| Technological advancement and innovation | Manned lunar flights drive innovations in life support, radiation protection, propulsion, and AI-robotics integration, tested in the harsh lunar environment. | 1. What human-in-the-loop testing protocols ensure reliable closed-loop life support systems for extended days? |
| Geopolitical and Strategic Leadership | Returning Humans to the Moon is promoting international partnerships while strategic outcomes include secured access to lunar resources. | 1. How do manned missions secure strategic lunar locations for sustained presence? |
| Economic and Commercial Opportunities | Human presence catalyses a lunar economy through mining, tourism and manufacturing with outcomes project a billion-dollar economy. | 1. How can manned missions demonstrate viable supply chains for off-Earth manufacturing? |

*3.3 Limitations and Challenges*

High radiation, extreme temperatures, and lunar regolith on the Moon pose risks to astronaut health and equipment. Uneven terrain complicates safe landings and precise positioning. Water ice at the poles creates challenges for sustainable resource development. Space robotics require high autonomy with limited resources for exploration, and human-robot cooperation is critical for safe operations. Human spaceflight, driven by socio-political factors like international cooperation and national prestige, offers diplomatic, economic, and inspirational benefits beyond research, making it a multifaceted endeavour.

4. **Conclusions**

The Moon represents a strategically imperative and technologically significant milestone in the progression of deep space exploration and the establishment of long-term human habitation beyond low Earth orbit. The integration of advanced Artificial Intelligence (AI) architectures and autonomous space robotic systems into lunar exploration frameworks is driving a transformative shift in mission autonomy, data processing, and operational efficiency. These intelligent systems enable enhanced scientific data acquisition, real-time decision-making, and precision-driven surface operations, thereby augmenting both robotic and crewed missions. Nevertheless, the deployment of such technologies in the lunar environment presents a multitude of complex challenges. These include extreme radiation exposure from galactic cosmic rays and solar particle events, abrasive and electrostatically charged lunar regolith that compromises mechanical and electronic systems, precision navigation in GPS-denied environments, communication latency and bandwidth limitations due to Earth-Moon distance constraints, and the assurance of fault-tolerant, high-reliability autonomous operations. To optimize mission outcomes while ensuring astronaut safety and system resilience, it is essential to adopt a hybrid exploration model that synergistically integrates human cognitive flexibility with AI-enabled robotic automation.